\documentclass[10pt,twocolumn,A4]{article}

\usepackage{cvpr}
\usepackage{times}
\usepackage{epsfig}
\usepackage{graphicx}
\usepackage{caption}
\usepackage{amsmath}
\usepackage{amssymb}
\usepackage{tabularx}
\usepackage[normalem]{ulem}
\usepackage{multirow}
\usepackage{xcolor}
\usepackage{abstract} 
\usepackage{xr}

\definecolor{road}{RGB}{128,64,128}
\definecolor{sidewalk}{RGB}{244,35,232}
\definecolor{building}{RGB}{70,70,70}
\definecolor{wall}{RGB}{102,102,156}
\definecolor{fence}{RGB}{190,153,153}
\definecolor{pole}{RGB}{153,153,153}
\definecolor{light}{RGB}{250,170,30}
\definecolor{sign}{RGB}{220,220,0}
\definecolor{vegetation}{RGB}{107,142,35}
\definecolor{terrain}{RGB}{152,251,152}
\definecolor{sky}{RGB}{70,130,180}
\definecolor{person}{RGB}{220,20,60}
\definecolor{rider}{RGB}{255,0,0}
\definecolor{car}{RGB}{0,0,142}
\definecolor{truck}{RGB}{0,0,70}
\definecolor{bus}{RGB}{0,60,100}
\definecolor{train}{RGB}{0,80,100}
\definecolor{motorcycle}{RGB}{0,0,230}
\definecolor{bicycle}{RGB}{119,11,32}

\newcommand{\makelegend}{\resizebox{\textwidth}{!}{\begin{tabular}{llllllllll}
\fcolorbox{road}{road}{\rule{0pt}{6pt}\rule{6pt}{0pt}} Road & 
\fcolorbox{sidewalk}{sidewalk}{\rule{0pt}{6pt}\rule{6pt}{0pt}} Sidewalk & 
\fcolorbox{building}{building}{\rule{0pt}{6pt}\rule{6pt}{0pt}} Building & 
\fcolorbox{wall}{wall}{\rule{0pt}{6pt}\rule{6pt}{0pt}} Wall & 
\fcolorbox{fence}{fence}{\rule{0pt}{6pt}\rule{6pt}{0pt}} Fence & \fcolorbox{pole}{pole}{\rule{0pt}{6pt}\rule{6pt}{0pt}} Pole & 
\fcolorbox{light}{light}{\rule{0pt}{6pt}\rule{6pt}{0pt}} Tr.Light & 
\fcolorbox{sign}{sign}{\rule{0pt}{6pt}\rule{6pt}{0pt}} Tr.Sign & 
\fcolorbox{vegetation}{vegetation}{\rule{0pt}{6pt}\rule{6pt}{0pt}} Vegetation & 
\fcolorbox{terrain}{terrain}{\rule{0pt}{6pt}\rule{6pt}{0pt}} Terrain \\ [1pt]
\fcolorbox{sky}{sky}{\rule{0pt}{6pt}\rule{6pt}{0pt}} Sky & 
\fcolorbox{person}{person}{\rule{0pt}{6pt}\rule{6pt}{0pt}} Person & 
\fcolorbox{rider}{rider}{\rule{0pt}{6pt}\rule{6pt}{0pt}} Rider & 
\fcolorbox{car}{car}{\rule{0pt}{6pt}\rule{6pt}{0pt}} Car & 
\fcolorbox{truck}{truck}{\rule{0pt}{6pt}\rule{6pt}{0pt}} Truck & 
\fcolorbox{bus}{bus}{\rule{0pt}{6pt}\rule{6pt}{0pt}} Bus & 
\fcolorbox{train}{train}{\rule{0pt}{6pt}\rule{6pt}{0pt}} Train & 
\fcolorbox{motorcycle}{motorcycle}{\rule{0pt}{6pt}\rule{6pt}{0pt}} Motorcycle & 
\fcolorbox{bicycle}{bicycle}{\rule{0pt}{6pt}\rule{6pt}{0pt}} Bicycle & 
\end{tabular}}}

\newcommand{\makelegenddfcn}{\resizebox{\textwidth}{!}{\begin{tabular}{llllllll}
\fcolorbox{road}{road}{\rule{0pt}{6pt}\rule{6pt}{0pt}} Road & 
\fcolorbox{sidewalk}{sidewalk}{\rule{0pt}{6pt}\rule{6pt}{0pt}} Sidewalk & 
\fcolorbox{building}{building}{\rule{0pt}{6pt}\rule{6pt}{0pt}} Building & 
\fcolorbox{pole}{pole}{\rule{0pt}{6pt}\rule{6pt}{0pt}} Pole & 
\fcolorbox{light}{light}{\rule{0pt}{6pt}\rule{6pt}{0pt}} Tr.Light & 
\fcolorbox{sign}{sign}{\rule{0pt}{6pt}\rule{6pt}{0pt}} Tr.Sign & 
\fcolorbox{vegetation}{vegetation}{\rule{0pt}{6pt}\rule{6pt}{0pt}} Vegetation & 
\fcolorbox{terrain}{terrain}{\rule{0pt}{6pt}\rule{6pt}{0pt}} Terrain \\ [1pt]
\fcolorbox{sky}{sky}{\rule{0pt}{6pt}\rule{6pt}{0pt}} Sky & 
\fcolorbox{person}{person}{\rule{0pt}{6pt}\rule{6pt}{0pt}} Person & 
\fcolorbox{rider}{rider}{\rule{0pt}{6pt}\rule{6pt}{0pt}} Rider & 
\fcolorbox{car}{car}{\rule{0pt}{6pt}\rule{6pt}{0pt}} Car & 
\fcolorbox{truck}{truck}{\rule{0pt}{6pt}\rule{6pt}{0pt}} Truck & 
\fcolorbox{bus}{bus}{\rule{0pt}{6pt}\rule{6pt}{0pt}} Bus & 
\fcolorbox{motorcycle}{motorcycle}{\rule{0pt}{6pt}\rule{6pt}{0pt}} Motorcycle & 
\fcolorbox{bicycle}{bicycle}{\rule{0pt}{6pt}\rule{6pt}{0pt}} Bicycle
\end{tabular}}}

\usepackage[normalem]{ulem}

\usepackage[pagebackref=true,breaklinks=true,letterpaper=true,colorlinks,bookmarks=false]{hyperref}

\cvprfinalcopy 

\title{Procedural Modeling and Physically Based Rendering for Synthetic Data Generation in Automotive Applications}

\author{Apostolia Tsirikoglou$^{1,}$\thanks{apostolia.tsirikoglou@liu.se}\quad
Joel Kronander$^{1}$\quad Magnus Wrenninge$^{2,}$\thanks{magnus@7dlabs.com}\quad Jonas Unger$^{1,}$\thanks{jonas.unger@liu.se}\\\vspace{-3mm}\quad\\$^{1}$Link\"oping University, Sweden \qquad $^{2}$7D Labs}

\begin{document}
\twocolumn[{%
\renewcommand\twocolumn[1][]{#1}%
\maketitle
\begin{center}
    \includegraphics[width=0.49\textwidth]{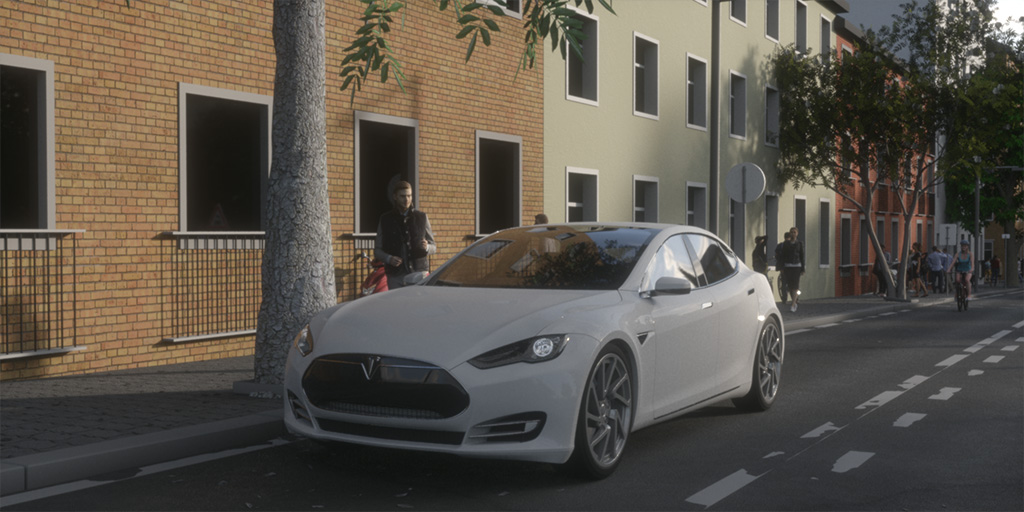}
    \includegraphics[width=0.49\textwidth]{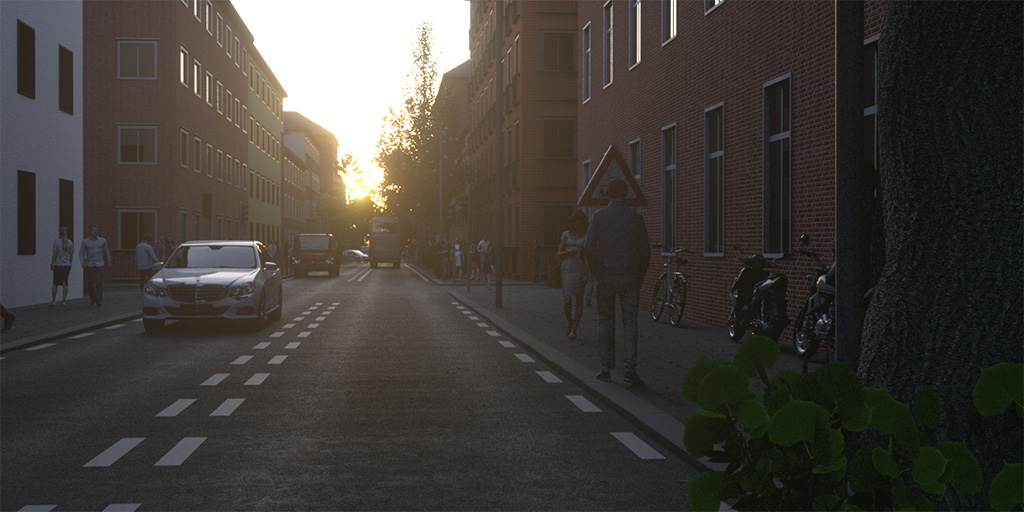}
    \captionof{figure}{Example images produced using our method for synthetic data generation. \label{fig:teaser}}
\end{center}%
}]

\saythanks


\begin{abstract}
\noindent We present an overview and evaluation of a new, systematic approach for generation of highly realistic, annotated synthetic data for training of deep neural networks in computer vision tasks. The main contribution is a procedural world modeling approach enabling high variability coupled with physically accurate image synthesis, and is a departure from the hand-modeled virtual worlds and approximate image synthesis methods used in real-time applications. The benefits of our approach include  flexible, physically accurate and scalable image synthesis, implicit wide coverage of classes and features, and complete data introspection for annotations, which all contribute to quality and cost efficiency. To evaluate our approach and the efficacy of the resulting data, we use semantic segmentation for autonomous vehicles and robotic navigation as the main application, and we train multiple deep learning architectures using synthetic data with and without fine tuning on organic (i.e. real-world) data. The evaluation shows that our approach improves the neural network's performance and that even modest implementation efforts produce state-of-the-art results. 
\end{abstract}

\section{Introduction}
\noindent Semantic segmentation is one of the most important methods for visual scene understanding, and constitutes one of the key challenges in a range of important applications such as autonomous driving, active safety systems and robot navigation. Recently, it has been shown that solutions based on deep neural networks \cite{AlexNet12, Simonyan15, Long_2015_CVPR} can solve this kind of computer vision task with high accuracy and performance. Although deep neural networks in many cases have proven to outperform traditional algorithms, their performance is limited by the training data used in the learning process. In this context, data itself has proven to be both the constraining and the driving factor of effective semantic scene understanding and object detection \cite{ILSVRC15, Simonyan15}. 

Access to large amounts of high quality data has the potential to accelerate the development of both new deep learning algorithms as well as tools for analyzing their convergence, error bounds, and performance. This has spurred the development of methods for producing synthetic, computer generated images with corresponding pixel-accurate annotations and labels. To date, the most widely used synthetic datasets for urban scene understanding are SYNTHIA \cite{Ros_2016_CVPR} and the dataset presented by Richter et al. \cite{Richter_2016_ECCV}. Both datasets use hand-modeled game worlds and rasterization-based image synthesis. It is worth noting that none of these previous studies have considered, in-depth, the way in which the virtual world itself is generated. Instead, focus has been put on the capture of existing 3D worlds and analyzing the performance of the resulting data. 

In recent years, game engines have steadily improved their ability to render realistic images. One recent example is the short film Adam\footnote{https://unity3d.com/pages/adam}, which was rendered in the Unity engine. However, it does not take long for a trained eye to spot the inconsistencies in these types of real-time renderings. In contrast, much of current visual effects in film feature imagery that even professionals cannot tell apart from reality. So far, current studies using synthetic data all involve mixing or fine tuning with organic datasets in order to achieve useful results. The \textit{domain shift} of the data is both discussed in these studies and obvious when viewing the images. Given that deep neural networks are reaching and sometimes exceeding human-level perception in computer vision tasks, it follows that synthetic data eventually needs to be as realistic as real data, if it is to become a useful complement, or at some point, a substitute. To that end, we avoid the use of the term "photo-realistic" throughout this paper; although our method is grounded in physically based image synthesis methods that can enable extremely realistic results, and although we achieve state-of-the-art results, neither ours nor the compared datasets can currently claim to be photo-realistic. Instead, we aim to make the reader conscious of the need to thoroughly analyze and evaluate realism in synthetic datasets. 
  
\begin{figure*}[!h]
{\centering
\includegraphics[width=1\linewidth]{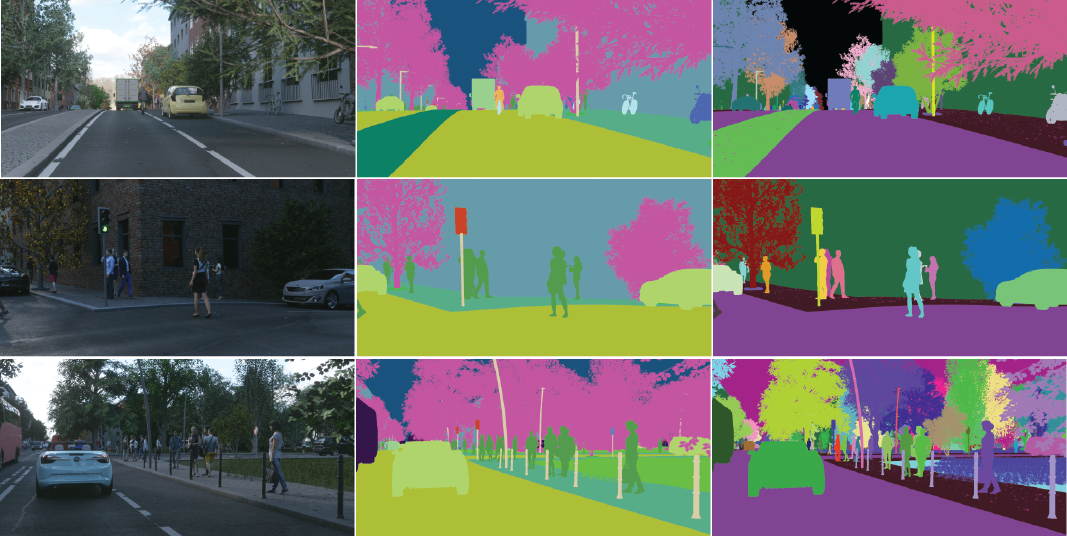}
\caption{\label{fig:example-images} \textbf{Left:} Three example images generated using our approach. \textbf{Middle:} The corresponding per-pixel class segmentation. \textbf{Right:} The instance segmentation. The procedural world modeling approach does not generate images from a fixed world, but rather samples from combinations of the input class instances described by the experimental design and scope. 
}
}
\end{figure*}

In this paper we use state-of-the-art computer graphics techniques, involving detailed geometry, physically based material representations, Monte Carlo-based light transport simulation as well as simulation of optics and sensors in order to produce realistic images with pixel-perfect ground truth annotations and labels, see Figures~\ref{fig:teaser} and~\ref{fig:example-images}. Our method combines procedural, automatic world generation with accurate light transport simulation and scalable, cloud-based computation capable of producing hundreds of thousands or millions of images with known class distributions and a rich set of annotation types. Compared to game engine pipelines, our method uses principles from the visual effects and film industries, where large scale production of images is well established and realism is paramount. 

Whereas image creation in film generally aims to produce a sequence of related images (i.e. an animation), we note that synthetic datasets instead benefit from images that are as diverse as possible. To that end, our system procedurally generates an entirely unique world for each output image from a set of classes representing vehicles, buildings, pedestrians, road surfaces, vegetation, trees and other relevant factors. All aspects of the individual classes, such as geometry, materials, color, and placement are parameterized, and a synthesized image and its corresponding annotations constitute a sampling of that parameter space. In the following sections, we demonstrate that this approach outperforms existing synthetic datasets in the semantic segmentation problem on multiple state-of-the art deep neural network architectures. 

\section{Background and Related work}
\label{sec:relatedwork}
\noindent The most common approach for producing training data with ground truth annotations has been to employ hand labeling, e.g.~CamVid~\cite{Brostow09}, Cityscapes \cite{Cordts2016Cityscapes}, the KITTI dataset~\cite{Geiger13}, or the Mapillary Vistas Dataset\footnote{www.mapillary.com}, which is both time consuming and complex to orchestrate. In Brostow et al.~\cite{Brostow09} and Cordts et al.~\cite{Cordts2016Cityscapes} it is reported that a single image may take from 20 to 90 minutes to annotate. Another  problem inherent to manual annotation is that some objects and features are difficult to classify and annotate correctly, especially when the image quality and lighting conditions vary. In the Cityscapes dataset, partially occluded objects, such as a pedestrian behind a car, are sometimes left unclassified, and the annotation of edges of dynamic objects, vegetation and foliage are often cursory.

Although several real-world, organic data sets are available, there has been a need to address the issue of data set bias~\cite{Torralba11,Khosla:2012:UDD:2402940.2402953}, by going beyond the thousands of hand labeled images they consist of, and in a controlled way ensuring a wider and generalizable coverage of features and classes within each training image. As analyzed in detail by Johnson et al.~\cite{Johnson-Roberson:2017aa}, game-based image synthesis has matured enough that the performance of deep learning architectures can be improved using computer generated footage with pixel accurate annotations. Synthetic data has been successfully used in a range of application domains including prediction of object pose \cite{Su_2015_ICCV, MovshovitzAttias2016, guptaCVPR15a}, optical flow \cite{Dosovitskiy15}, semantic segmentation for indoor scenes \cite{Handa16,Zhang_2017_ICCV}, and analysis of image features \cite{Aubry15, Kaneva_2011}. 

Previous methods for data generation in automotive applications have largely used computer game engines. Richter et al.~\cite{Richter_2016_ECCV} and Johnson et al.~\cite{Johnson-Roberson:2017aa} used the Grand Theft Auto (GTA) engine from Rockstar Games, and Ros et al.~\cite{Ros_2016_CVPR} and Gaidon et al.~\cite{cvpr16_virtual_worlds} used the Unity development platform\footnote{www.unity3d.com} for the SYNTHIA and Virtual KITTI data set respectively. Other examples of using video game engines are presented by Shafaei et al. \cite{ShafaeiLS16} who are using an undisclosed game engine, and Qui et al.~\cite{QiuY16} who recently introduced the UnrealCV plugin, which enables generation of image- and accompanying ground truth annotations using Unreal Engine\footnote{www.unrealengine.com} from Epic Games. Although it provides relatively easy access to virtual worlds, the game engine approach to synthetic data generation is limited in several ways. First, pre-computations with significant approximations of the light transport in the scene are required in order to fit the world model onto the GPU and enable interactive rendering speeds. Consequently, these methods do not scale well when the ratio of final images to number of scenarios (i.e. game levels) is low. Secondly, even though the 3D world in many cases may be large, it is hand modeled and of finite size. This not only means that it is time consuming and costly to build, but also that  the coverage of classes and features is limited, and that dataset bias is inevitable. This is obvious if we consider the limit case: the more images we produce, the more similar each image becomes to the others in the dataset. Conversely, if a large number of scenarios were built, the cost of pre-computing light transport within each scene would not be amortized over a large enough set of images to be efficient. Finally, some important aspects such as accurate simulation of sensor characteristics, optics and surface and volumetric scattering (and even some annotations) may be impractical to include in rasterization-based image synthesis. 

Another approach for increasing variability and coverage in organic, annotated footage is to augment the images with synthetic models. Rozantsev et al.~\cite{Rozantsev:2015:RSI:2798735.2798831} proposed a technique where 3D models were superimposed onto real backgrounds through simulation of the camera system and lighting conditions in the real scene for improving the detection of aerial drones and aircrafts. Other interesting examples of data augmentation using image synthesis include pedestrian detection presented in Marin et al.~\cite{MVG2010}, the GAN-based gaze and hand pose detection described by Shrivastava et al.~\cite{shrivastava2016learning}, and rendering of 3D car models into background footage for segmentation tasks described by Alhaija et al.~\cite{Alhaija2017BMVC}. Although a smaller set of hand-annotated images can be augmented to include a wider variation, this approach is best suited for coarse annotations and does not generalize well to large image volumes and semantic segmentation tasks.

In contrast to previous methods for generation of annotated synthetic training data, we employ procedural world modeling (for an overview, see Ebert et al.~\cite{Ebert98}). The benefit is that the creation of the 3D world can be parameterized to ensure detailed control over class and feature variations. In our system, the user input is no longer a large, concrete 3D world, but rather a composition of classes and a scenario scope, from which the system creates only what is visible from each image's camera view. For image synthesis, we use physically based rendering techniques based on path tracing and Monte Carlo integration~\cite{Kajiya:1986:RE:15886.15902}. This allows accurate simulation of sensors, optics, and the interaction between the light, materials and geometry in the scene. The next section gives an overview of our system and the underlying techniques used to generate synthetic training data.

\section{Method}
\label{sec:dataset}
\begin{figure*}[!h]
{\centering
\includegraphics[width=1\linewidth]{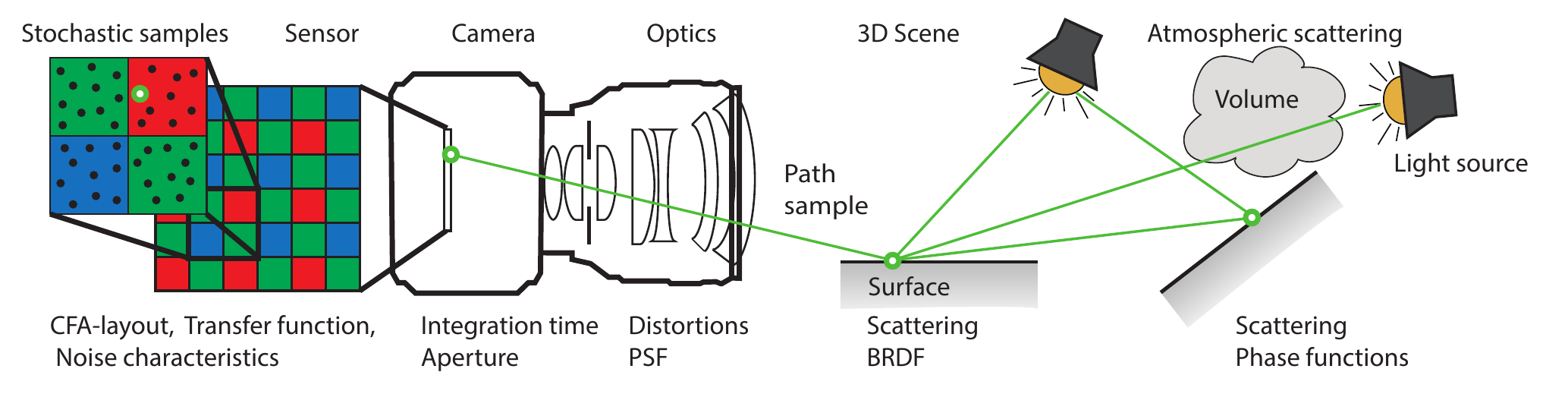}
{\caption{\label{fig:lighttransport} Illustration of the principles of path tracing. Sample paths are stochastically generated in the image plane and traced through the scene. At each interaction the light emitting objects are sampled and their contribution summed up. The technique enables accurate simulation of sensor characteristics and the color filter array (CFA), the effect of the optical system (PSF, distortion, etc.), complex geometries and  scattering at surface and in participating media. Path tracing is computationally expensive, but parallelizes and scales well, and is made feasible through Monte Carlo importance sampling techniques.}}
}
\end{figure*}
\noindent Image synthesis requires a well-defined model of the 3D virtual scene. The model contains the geometric description of objects in the scene, a set of materials describing the appearance of the objects, specifications of the light sources in the scene, and a virtual camera model. The geometry is often specified in terms of discretized rendering primitives such as triangles and surface patches. The materials define how light interacts with surfaces and participating media. Finally, the scene is illuminated using one or several light sources, and the composition of the rendered frame is defined by introducing a virtual camera. 

One way to consider realism in synthetic imagery is as a multiplicative quantity. That is, in order to produce realistic synthetic images we must ensure that each of steps involved are capable of producing realistic results. For example, there is little hope of achieving  realistic imagery even when employing physically accurate light transport if the geometry is of poor quality, and vice versa. To this end, we identify five orthogonal aspects of realism that are key to achieving the goal:

\begin{enumerate}
\setlength\itemsep{-0.2em}
\item Overall scene composition
\item Geometric structure
\item Illumination by light sources
\item Material properties
\item Optical effects
\end{enumerate}

Our method addresses each of these aspects. Realism in the overall scene composition along with the geometric structure and the material properties is addressed by the rule set used in the procedural modeling process. Realism in the illumination and material interaction is addressed by physically based light transport simulation and optical effects are modeled using point spread functions in the image domain.

\subsection{Physically based light transport simulation}
\noindent The transfer of light from light sources to the camera, via surfaces and participating media in the scene, is described by light transport theory, which is a form of radiative transfer~\cite{chandrasekhar1960radiative}. For surfaces, the light transport is often described using the macroscopic geometric-optics model defined by the rendering equation~\cite{Kajiya:1986:RE:15886.15902}, expressing the outgoing radiance $L(\vec{x} \rightarrow \omega_o)$ from a point $\vec{x}$ in direction $\omega_o$ as

\begin{align}
\label{eq:renderingequation}
L(\vec{x} \rightarrow \omega_o) =& L_e(\vec{x} \rightarrow \omega_o) + \\\nonumber
&\underbrace{\int_{\Omega}{L(\vec{x} \leftarrow \omega_i)\rho(\vec{x},\omega_i,\omega_o)(\vec{n}\cdot\omega_o) d\omega_i}}_{L_r(\vec{x} \rightarrow \omega_o)},
\end{align}

\noindent where $L(\vec{x} \leftarrow \omega_i)$ is the incident radiance arriving at the point $\vec{x}$ from direction $\omega_i$, $L_e(\vec{x} \rightarrow \omega_o)$ is the radiance emitted from the surface, $L_r(\vec{x} \rightarrow \omega_o)$ is the reflected radiance, $\rho(\vec{x},\omega_i,\omega_o)$ is the \textit{bidirectional reflectance distribution function} (BRDF) describing the reflectance between incident and outgoing directions~\cite{Nicodemus:65}, $\Omega$ is the visible hemisphere, and $\vec{n}$ is the surface normal at point $\vec{x}$. 

Rendering is carried out by simulating the light reaching the camera. This requires solving the rendering equation for a large number of sample points in the image plane which is a set of potentially millions of inter-dependent, high-dimensional analytically intractable integral equations. Solving the rendering equation is challenging since the radiance $L$ appears both inside the integral expression and as the quantity we are solving for. The reason for this is that the outgoing radiance at any one point affects the incident radiance at all other points in the scene. This results in a very large system of nested integrals. Formally, the rendering equation is a Fredholm integral equation of the second kind, for which analytic solutions are impossible to find in all but the most trivial cases. In practice, the rendering problem can be solved in a number of ways with different light transport modeling techniques, the most common of which can be divided into two main categories: \emph{rasterization} which is the method generally used in GPU rendering, and \emph{path tracing} in which equation~\ref{eq:renderingequation} is solved using Monte Carlo integration techniques which stochastically construct paths that connect light sources to the virtual camera and compute the path energy throughput as illustrated in Figure~\ref{fig:lighttransport}.  

The rendering system used in this paper relies on path tracing as it is the most general rendering algorithm. In the limit, path tracing can simulate any type of lighting effects including multiple light bounces and combinations between complex geometries and material scattering behaviors. Another benefit is that it is possible to sample the scene being rendered over both the spatial and temporal dimensions. For example, by generating several path samples per pixel in the virtual film plane it is possible to simulate the area sampling over the extent of each pixel on a real camera sensor, which in practice leads to efficient anti-aliasing in the image and enables simulation of the point spread function (PSF) introduced by the optical system. By distributing the path samples over time by transforming (e.g. animating) the virtual camera and/or objects in the scene, it is straightforward to accurately simulate motion blur, which is a highly important aspect in the simulation of the computer vision system on a vehicle. Path tracing is a standard tool in film production and is implemented in many rendering engines. For an in-depth introduction to path tracing and the plethora of techniques for reducing the computational complexity such as Monte Carlo importance sampling and efficient data structures for accelerating the geometric computations, we refer the reader to the textbook by Pharr et al.~\cite{Pharr:2010:PBR:1854996}.

\subsection{World generation using procedural modeling}
\label{sec:procedural}

\noindent An important notion in machine learning in general, and in deep learning in particular, is that of \textit{factors of variation}. In their textbook, Goodfellow et al.~\cite{Goodfellow16} state in their introduction that "Such factors are often not quantities that are directly observed. Instead, they may exist as either unobserved objects or unobserved forces in the physical world that affect observable quantities." (pp. 4-5). This notion is directly parallel to a longstanding methodology often employed in art and film production, namely procedural modeling. In designing our method for creating virtual worlds, we consider the procedural modeling aspect to be a direct path towards producing datasets with precisely controlled factors of variation.

Previous methods for producing synthetic datasets generally employ a single virtual world in which a virtual camera is moved around. However, this approach yields datasets where some environmental parameters are varied but others are constant or only partially varied throughout the world. Ideally, all parameters would be varied such that each image can be made as different from others as possible. Unfortunately, the architecture of game engines do not lend themselves well to this type of wide variability due to the divergence between the needs of game play and those of synthetic dataset production.

In order to produce a highly varied dataset, our method instantiates an entirely unique virtual world for each image. Although more time consuming than re-using the same geometric construct for multiple images, it is made practical by generating only the set of geometry that is visible either directly to the camera, or through reflections and shadows cast into the view of the camera.

When constructing the virtual world, we define a set of parameters to vary as well as a \textit{rule set} that translates the parameter values into the concrete scene definition. It is worth noting that this approach yields an exponential explosion of potential worlds. However, in the context of dataset generation, this is entirely beneficial and means that each added factor of variation multiplies rather than adds to the size of the parameter space.

The following list highlights some of the key procedural parameters used in producing the dataset:

\begin{itemize}
\setlength\itemsep{-0.2em}
\item \textbf{Road} width; number of lanes; material; repair marks; cracks
\item \textbf{Sidewalk} width; curb height; material; dirt amount
\item \textbf{Building} height and width; window height, width and depth;  material
\item \textbf{Car} type; count; placement; color
\item \textbf{Pedestrian} model; count; placement (in road, on sidewalk)
\item \textbf{Vegetation} type; count; placement
\item \textbf{Sun} longitude; latitude
\item \textbf{Cloud cover} amount
\item \textbf{Misc.} Placement and count of poles, traffic lights, traffic signs, etc.

\end{itemize}

Our virtual world uses a mixture of both procedurally generated geometry, as well as model libraries. For example, the buildings, road surface, sidewalks, traffic lights and poles are all procedurally generated and individually unique. For pedestrians, bicyclists, cars and traffic signs, we use model libraries, where the geometry is shared between all instances, but properties such as placement, orientation and certain texture and material aspects vary between instances. Despite using a small set of prototypes for these classes, the resulting dataset is still rich, due to the large variability in how these classes are seen. In addition, the rule set used to populate the virtual world includes expected contextual arrangements such as pedestrians in cross walks, cars and bicyclists crossing the road, etc.

The illumination of the scene is specified by a sun position and includes an accurate depiction of the sky, including cloud cover. This ensures that the lighting conditions at street level includes a continuous range of times of day, all potential light directions relative to the ego vehicle, as well as indirect light due to clouds and other participating media. Illumination is calculated in a high dynamic range, scene-referred linear RGB color space, ensuring that the virtual camera sees realistic light and contrast conditions.

Overall, we have chosen to focus the development effort for the evaluation dataset on the most relevant classes for automotive purposes: road, sidewalk, traffic light and sign, person, rider, car and bicycle. These classes exhibit greater model variation, geometric detail, material fidelity and combinatoric complexity compared to the other classes, which are present but less refined. The evaluations presented in the next section show that the higher realism and feature variation in the selected classes increases the accuracy of the semantic segmentation.

\section{Evaluation}
\label{sec:results}

\begin{table*}[h]
\centering
\resizebox{\textwidth}{!}{\begin{tabular}{l|l||l|l|l|l|l|l|l|l|l|l|l|l|l|l|l|l||l}
               & Dataset & \rotatebox[origin=c]{90}{Road}   &  \rotatebox[origin=c]{90}{Sidewalk} &  \rotatebox[origin=c]{90}{Building} &  \rotatebox[origin=c]{90}{Pole}   &  \rotatebox[origin=c]{90}{Tr.Light} &  \rotatebox[origin=c]{90}{Tr.Sign} &  \rotatebox[origin=c]{90}{Vegetation} &  \rotatebox[origin=c]{90}{Terrain} &  \rotatebox[origin=c]{90}{Sky}    &  \rotatebox[origin=c]{90}{Person} &  \rotatebox[origin=c]{90}{Rider}  &  \rotatebox[origin=c]{90}{Car}    &  \rotatebox[origin=c]{90}{Truck}  &  \rotatebox[origin=c]{90}{Bus}    &  \rotatebox[origin=c]{90}{Motorcycle} &  \rotatebox[origin=c]{90}{Bicycle} &  \rotatebox[origin=c]{90}{Mean IoU} \\[20pt]
               \hline 
               
\multirow{7}{*}{DFCN -- frontend}& S & 0.44 & 18.54 & 35.25 & 15.00 & 0.00 & 0.00 & 64.91 & 0.00 & 72.09 & 49.34 & 2.81  & 60.51 & 0.00 & \textbf{11.47} & 0.06 & 0.83   & 20.7 \\

& GTA & 54.82 & 21.82 & \textbf{66.37} & 18.01 & 11.89 & 4.31 & \textbf{79.02} & \textbf{30.43} & \textbf{72.99} & 40.56 & 2.39  & \textbf{73.49} & \textbf{11.33} & 8.58 & 1.84 & 0.00 & 31.12 \\

& O & \textbf{71.33} & \textbf{34.29}   & 63.33   & \textbf{33.33} & \textbf{23.24}        & \textbf{28.33}       & 72.58     & 5.99   & 67.22 & \textbf{49.67} & \textbf{26.21} & 50.97 & 7.10  & 5.19  & \textbf{3.14}      & \textbf{48.89}  & \textbf{36.93}   \\ 
\cline{2-19}

& CS & 96.41 & 75.97   & 90.04   & 50.61 & 50.16        & 65.17       & 90.67     & 54.31  & 90.85 & 72.34 & 44.69 & 89.99 & 40.62 & 59.08 & 43.81     & 68.57  & 67.71 \\

& S + CS & 96.46 & 75.77 & 90.10 & 50.13 & 49.65 & 64.81 & 90.51 & 55.72 & 91.22 & 73.41 & \textbf{45.43} & 90.28 & 45.78 & \textbf{66.03} & 47.06 & 68.87 & 68.83 \\

& GTA + CS & 96.62 & 76.97 & \textbf{90.34} & \textbf{51.08} & \textbf{51.47} & 64.86 & \textbf{90.96} & \textbf{58.38} & 91.02 & 73.44 & 44.24 & 90.59 & 45.75 & 63.16 & 46.59 & 69.00 & 69.03 \\

& O + CS & \textbf{96.70} & \textbf{77.26} & 90.30 & 49.58 & 51.34 & \textbf{65.24} & 90.85 & 57.34 & \textbf{91.34} & \textbf{74.30} & 45.25 & \textbf{90.95} & \textbf{47.24} & 65.69 & \textbf{47.75} & \textbf{70.09} & \textbf{69.45} \\ 

\hline

\multirow{4}{*}{DFCN -- context} &
CS & 96.48 & 76.96 & 90.24 & 51.16 & 51.18 & 66.99 & 90.98 & 57.83 & 91.74 & 71.76 & 46.13 & 90.04 & 47.00 & 65.52 & 45.02 & 64.76 & 68.97 \\ 

& S & 0.46 & 17.66 & 34.39 & 13.24 & 0.00 & 0.11 & 63.83 & 0.00 & 72.26 & 47.64 & 2.58 & 61.34 & 0.00 & 7.90 & 0.64 & 1.30 & 20.21 \\ 
& GTA & 54.07 & 22.10 & 60.95 & 20.25 & 20.52 & 4.54 & \textbf{78.98} & \textbf{26.57} & \textbf{72.38} & 43.95 & 1.10 & 69.19 & \textbf{15.06} & \textbf{12.83} & \textbf{6.92} & 0.00 & 31.84 \\ 
& O & \textbf{74.42} & \textbf{37.85} & \textbf{70.26} & \textbf{32.91} & \textbf{27.32} & \textbf{28.85} & 63.89 & 9.89 & 70.44 & \textbf{52.76} & \textbf{26.23} & \textbf{73.73} & 10.22 & 9.21 & 3.30 & \textbf{45.50} & \textbf{39.80} \\ 

\hline

\end{tabular}}
\caption{DFCN \cite{YuKoltun2016} results on Cityscapes \cite{Cordts2016Cityscapes} (CS), SYNTHIA \cite{Ros_2016_CVPR} (S), Richter et al. \cite{Richter_2016_ECCV} (GTA) and Our data (O). 
}
\label{table:dfcn_results}
\end{table*}

\begin{table*}[h]
\centering
\resizebox{\textwidth}{!}{\begin{tabular}{l|l||l|l|l|l|l|l|l|l|l|l|l|l|l|l|l|l}
 & Dataset & 
 \rotatebox[origin=c]{90}{Road}   &  \rotatebox[origin=c]{90}{Sidewalk} &  \rotatebox[origin=c]{90}{Building} &  \rotatebox[origin=c]{90}{Pole}   &  \rotatebox[origin=c]{90}{Tr.Light} &  \rotatebox[origin=c]{90}{Tr.Sign} &  \rotatebox[origin=c]{90}{Vegetation} &  \rotatebox[origin=c]{90}{Terrain} &  \rotatebox[origin=c]{90}{Sky}    &  \rotatebox[origin=c]{90}{Person} &  \rotatebox[origin=c]{90}{Rider}  &  \rotatebox[origin=c]{90}{Car}    &  \rotatebox[origin=c]{90}{Truck}  &  \rotatebox[origin=c]{90}{Bus}    &  \rotatebox[origin=c]{90}{Motorcycle} &  \rotatebox[origin=c]{90}{Bicycle} \\[20pt] \hline 
\multirow{7}{*}{DFCN -- frontend}
& S & 1.04 & 18.54 & 36.25 & 17.37 & 0.00 & 1.71 & 67.15 & 0.00 & 72.41 & 49.55 & 3.90 & 62.42 & 0.00 & 11.47 & 0.32 & 0.83 \\
& GTA & 63.45 & 24.32 & \textbf{68.51} & 18.54 & 16.22 & 5.98 & \textbf{81.12} & \textbf{30.43} & \textbf{72.99} & 41.59 & 3.68 & \textbf{73.49} & \textbf{15.02} & \textbf{13.01} & 2.47 & 0.00 \\
& O & \textbf{78.42} & \textbf{38.21} & 64.02 & \textbf{33.44} & \textbf{24.61} & \textbf{32.57} & 77.09 & 9.06 & 72.35 & \textbf{51.98} & \textbf{26.93} & 57.55 & 9.23 & 5.83 & \textbf{3.14} & \textbf{49.29} \\ 
\cline{2-18}

& CS & 96.56 & 76.76 & 90.12 & 50.61 & 53.38 & 68.16 & 90.85 & 58.72 & 91.29 & 73.98 & 44.69 & 90.12 & 40.62 & 60.78 & 44.38 & 70.72 \\ 

& S + CS & 96.56 & 76.68 & 90.20 & 50.35 & 49.94 & 64.87 & 90.82 & 57.26 & 91.30 & 73.61 & 45.45 & 90.29 & 45.78 & 66.03 & \textbf{48.56} & 69.02 \\

& GTA + CS & 96.62 & 76.97 & 90.35 & \textbf{51.08} & 51.66 & 65.19 & \textbf{91.11} & 58.49 & \textbf{91.64} & 73.44 & 44.81 & 90.59 & 47.21 & 64.78 & 48.14 & 69.47 \\

& O + CS & \textbf{96.72} & \textbf{77.26} & \textbf{90.45} & 51.06 & \textbf{52.22} & \textbf{66.09} & 90.98 & \textbf{58.87} & 91.54 & \textbf{74.59} & \textbf{47.27} & \textbf{90.96} & \textbf{47.64} & \textbf{66.70} & 47.97 & \textbf{70.23}  \\ \hline
\multirow{4}{*}{DFCN -- context}

& CS & 96.53 & 77.24 & 90.41 & 51.50 & 51.99 & 67.18 & 91.08 & 57.99 & 91.78 & 72.43 & 46.84 & 90.36 & 47.00 & 65.98 & 46.10 & 67.22 \\ 
& S & 0.48 & 17.73 & 34.52 & 13.36 & 0.00 & 0.24 & 64.01 & 0.00 & 72.30 & 47.66 & 2.63 & 61.36 & 0.00 &  8.44 & 0.65 & 1.35 \\
& GTA & 60.28 & 22.98 & 62.42 & 20.55 & 20.61 & 4.59 & \textbf{78.98} & \textbf{26.57} & \textbf{72.47} & 43.95 & 1.17 & 69.19 & \textbf{15.06} & \textbf{12.83} & \textbf{7.05} & 0.00 \\
& O & \textbf{76.04} & \textbf{38.52} & \textbf{70.38} & \textbf{32.96} & \textbf{27.66} & \textbf{28.89} & 67.15 & 9.93 & 70.46 & \textbf{53.01} & \textbf{26.81} & \textbf{73.73} & 10.23 & 9.28 & 3.39 & \textbf{46.57} \\ 
\hline

\end{tabular}}
\caption{Best per-class IoU for all validation iterations on Cityscapes \cite{Cordts2016Cityscapes} (CS), SYNTHIA \cite{Ros_2016_CVPR} (S), Richter et al. \cite{Richter_2016_ECCV} (GTA) and Our data (O) for DFCN \cite{YuKoltun2016} architecture.
}
\label{table:dfcn_best_class_IoU}
\end{table*}

\begin{table*}[h]
\centering
\resizebox{\textwidth}{!}{\begin{tabular}{l||l|l|l|l|l|l|l|l|l|l|l|l|l|l|l|l|l|l|l||l}
             & \rotatebox[origin=c]{90}{Road}   &  \rotatebox[origin=c]{90}{Sidewalk} &  \rotatebox[origin=c]{90}{Building} &  \rotatebox[origin=c]{90}{Wall}   &  \rotatebox[origin=c]{90}{Fence}   &  \rotatebox[origin=c]{90}{Pole}   &  \rotatebox[origin=c]{90}{Tr.Light} &  \rotatebox[origin=c]{90}{Tr.Sign} &  \rotatebox[origin=c]{90}{Vegetation} &  \rotatebox[origin=c]{90}{Terrain} &  \rotatebox[origin=c]{90}{Sky}    &  \rotatebox[origin=c]{90}{Person} &  \rotatebox[origin=c]{90}{Rider}  &  \rotatebox[origin=c]{90}{Car}    &  \rotatebox[origin=c]{90}{Truck}  &  \rotatebox[origin=c]{90}{Bus}    &  \rotatebox[origin=c]{90}{Train} &  \rotatebox[origin=c]{90}{Motorcycle} &  \rotatebox[origin=c]{90}{Bicycle} &  \rotatebox[origin=c]{90}{Mean IoU} \\[20pt]
               \hline 
               
S & 60.77 & 28.04 & 59.75 & 0.07 & 0.07 & 25.63 & 2.34 & 2.69 & 74.59 & 0.00 & 74.83 & 38.35 & 3.84 & 35.56 & 0.00 & 2.09 & 0.00 & \textbf{1.92} & 2.74 & 21.75 \\
GTA & 40.30 & 21.20 & 62.45 & \textbf{7.17} & \textbf{6.85} & 0.00  & \textbf{11.03} & 1.52 & \textbf{75.40} & 12.59 & 59.83 & 31.72 & 0.00 & 27.30 & \textbf{14.91} & \textbf{7.47} & \textbf{7.98} & 0.23 & 0.02 & 20.42 \\
O & \textbf{85.84} & \textbf{44.45} & \textbf{67.05} & -- & -- & \textbf{29.34} & 10.50 & \textbf{24.45} & 70.09 & \textbf{13.51} & \textbf{80.10} & \textbf{50.67} & \textbf{20.25} & \textbf{60.51} & 5.68  & 7.41 & -- & 1.18 & \textbf{20.91} & \textbf{31.15} \\

\hline

CS & 97.49 & 80.43 & 90.41 & 41.47 & 43.77 & 61.39 & 61.95 & 71.58 & 91.34 & \textbf{61.86} & \textbf{94.04} & 75.11 & 51.36 & 92.90 & 56.56 & 64.52 & 46.59 & 42.62 & 69.53 & 68.15 \\
S + CS         & \textbf{97.58} & \textbf{81.04} & \textbf{90.81} & 47.58 & \textbf{50.49} & 62.48 & 63.05 & 73.45 & 91.47 & 60.39 & 93.8  & 77.11 & 53.05 & 93.19 & 57.04 & 73.21 & 52.64 & 38.07 & 71.51 & 69.89 \\
GTA + CS       & 96.90 & 77.17 & 90.71 & \textbf{49.20} & 48.62 & 62.42 & 61.58 & 72.34 & 91.25 & 60.93 & 93.84 & 75.53 & 53.77 & \textbf{93.64} & \textbf{64.19} & 73.13 & \textbf{61.44} & \textbf{46.80} & 70.96 & 70.76 \\
O + CS         & 97.36 & 80.77 & 90.80 & 45.95 & 48.21 & \textbf{63.57} & \textbf{64.73} & \textbf{76.16} & \textbf{91.60} & 60.59 & 93.69 & \textbf{77.41} & \textbf{55.36} & 93.57 & 62.30 & \textbf{74.43} & 55.72 & 46.01 & \textbf{71.93} & \textbf{71.06} \\

\hline

\end{tabular}}
\caption{FRRN-A \cite{pohlen2017FRRN} results on Cityscapes \cite{Cordts2016Cityscapes} (CS), SYNTHIA \cite{Ros_2016_CVPR} (S), Richter et al. \cite{Richter_2016_ECCV} (GTA) and Our data (O).}
\label{table:frrn_results}
\end{table*}

\begin{table*}[h]
\centering
\resizebox{\textwidth}{!}{\begin{tabular}{l||l|l|l|l|l|l|l|l|l|l|l|l|l|l|l|l|l|l|l}
               & \rotatebox[origin=c]{90}{Road}   &  \rotatebox[origin=c]{90}{Sidewalk} &  \rotatebox[origin=c]{90}{Building} &  \rotatebox[origin=c]{90}{Wall}   &  \rotatebox[origin=c]{90}{Fence}   &  \rotatebox[origin=c]{90}{Pole}   &  \rotatebox[origin=c]{90}{Tr.Light} &  \rotatebox[origin=c]{90}{Tr.Sign} &  \rotatebox[origin=c]{90}{Vegetation} &  \rotatebox[origin=c]{90}{Terrain} &  \rotatebox[origin=c]{90}{Sky}    &  \rotatebox[origin=c]{90}{Person} &  \rotatebox[origin=c]{90}{Rider}  &  \rotatebox[origin=c]{90}{Car}    &  \rotatebox[origin=c]{90}{Truck}  &  \rotatebox[origin=c]{90}{Bus}    &  \rotatebox[origin=c]{90}{Train} &  \rotatebox[origin=c]{90}{Motorcycle} &  \rotatebox[origin=c]{90}{Bicycle} \\[20pt]
               \hline 
S  & 68.13 & 29.01 & 68.29 & 4.03 & 0.28 & 26.28 & 4.55 & 5.60 & 76.71 & 0.00 & 82.00 & 40.80 & 7.17 & 36.86 & 0.00 & 5.82 & 0.00 & 3.44 & 9.37 \\ 
GTA  & 60.07 & 21.20 & \textbf{70.11} & \textbf{8.80} & \textbf{15.79} & 0.0 & 12.15 & 3.89 & \textbf{77.40} & \textbf{19.65} & 77.92 & 37.74 & 3.10 & 45.80 & \textbf{20.29} & \textbf{16.79} & \textbf{17.50} & 1.92 & 1.07 \\ 
O  & \textbf{90.87} & \textbf{47.36} & 69.31 & -- & -- & \textbf{33.26} & \textbf{13.58} & \textbf{26.84} & 75.58 & 15.67 & \textbf{83.19} & \textbf{54.38} & \textbf{26.10} & \textbf{70.18} & 19.27 & 13.07 & -- & \textbf{4.81} & \textbf{28.31} \\ \hline

CS & 97.65 & 81.49 & 90.61 & 48.56 & 46.69 & 61.92 & 62.42 & 72.33 & 91.43 & 61.86 & 94.30 & 76.08 & 53.55 & 93.46 & 57.97 & 69.56 & 50.71 & 47.16 & 70.47 \\
S + CS      & 97.73 & 82.10 & 90.98 & 50.98 & \textbf{52.23} & 63.95 & 65.01 & 74.44 & 91.82 & 62.72 & 94.46  & 78.00 & 56.54 & 93.35 & 60.82 & 75.09 & 60.95 & 44.80 & 72.18 \\ 
GTA + CS     & 97.73 & 81.57 & 91.03 & \textbf{54.77} & 50.54 & 63.00 & 64.03 & 73.72 & 91.66 & 61.18 & \textbf{94.49} & 76.75 & 56.58 & \textbf{93.64} & 65.59 & 76.50 & \textbf{63.59} & 47.90 & 71.65 \\ 
O + CS     & \textbf{97.82} & \textbf{82.25} & \textbf{91.08} & 54.68 & 51.01 & \textbf{64.50} & \textbf{65.83} & \textbf{76.66} & \textbf{91.92} & \textbf{63.14} & 94.46 & \textbf{78.20} & \textbf{57.28} & 93.58 & \textbf{66.10} & \textbf{77.38} & 62.31 & \textbf{48.51} & \textbf{72.74}

\end{tabular}}
\caption{
Best per-class IoU for all validation iterations on Cityscapes \cite{Cordts2016Cityscapes} (CS), SYNTHIA \cite{Ros_2016_CVPR} (S), Richter et al. \cite{Richter_2016_ECCV} (GTA) and Our data (O) for FRRN-A \cite{pohlen2017FRRN} architecture.}
\label{table:best_class_iou}
\end{table*}

\externaldocument{results.tex}

\newlength{\longword}
\settowidth{\longword}{FRRN-A}

\noindent To evaluate the proposed data synthesis pipeline and the resulting dataset, we perform a series of experiments wherein we train different state-of-the-art neural network architectures using synthetic data, and combinations of synthetic and organic data from Cityscapes. We compare the performance of our dataset to the two well-known synthetic datasets by Ros et al. (SYNTHIA) \cite{Ros_2016_CVPR}, and Richter et al. \cite{Richter_2016_ECCV}. Although it is difficult to say exactly how much time was spent producing these two virtual worlds, it is worthwhile to remember that one was produced by a research lab while the other represents the direct and indirect work of over 1,000 people.

\subsection{Methodology}

\begin{figure}[t]
{
\includegraphics[width=1\columnwidth, trim={4cm 0 7cm 0},clip]{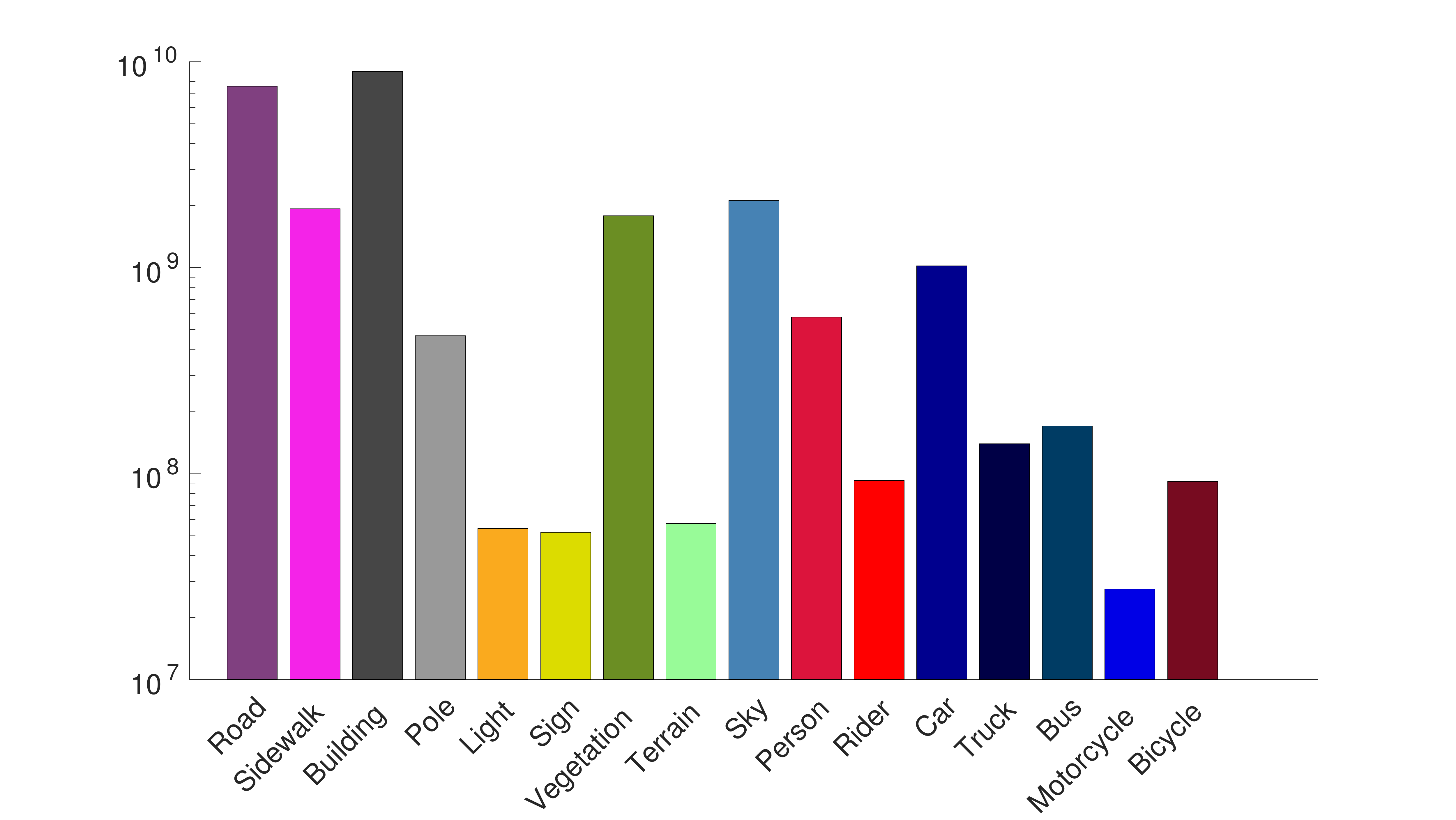}
\caption{Number of annotated pixels per class in our dataset.}
\label{fig:anotated_pixels_per_class}
}
\end{figure}

\noindent 
We use semantic segmentation as the benchmark application, and compare the performance obtained when the network is trained using our synthetic data, consisting of 25,000 images, to that obtained using SYNTHIA, with 9,400 training images, and the set of 20,078 images\footnote{The DFCN and FRRN architectures require consistent resolution for all inputs, so we use the subset of the original 25,000 images that have identical resolution.} presented by Richter et al. Figure \ref{fig:anotated_pixels_per_class} shows the distribution of classes in our dataset. We evaluate the performance of both the synthetic data alone, and by subsequent fine-tuning on the Cityscapes training set. The network performance is computed from inference results on the Cityscapes validation set and quantified using the intersection over union (IoU) metric. For the comparison we choose two architectures, and use the publicly available reference implementations without any further modifications:

\begin{enumerate}
	\item \phantomsection \label{dfcn} \textbf{DFCN} -- A dilated fully convolutional network, as presented by Yu and Koltun \cite{YuKoltun2016}, which proposes an exponential schedule of dilated convolutional layers as a way to combine local and global knowledge. This architecture integrates information from different spatial scales and balances local, pixel-level accuracy, e.g. precise detection of edges, and knowledge of the wider, global level. The architecture consists of the frontend module along with a context aggregation module, where several layers of dilation can be applied. 
    
    The DFCN-frontend network was trained using stochastic gradient descent with a learning rate of $10^{-5}$, momentum of $0.99$ and a batch size of $8$ for synthetic data. For organic data,
we used a learning rate of $10^{-4}$ in baseline training and $10^{-5}$ in fine-tuning, with the same momentum and batch size. For both frontend baseline and fine-tuning trainings for all datasets each crop is of size $628\times628$. The DFCN-context network was also trained using stochastic gradient descent with a learning rate of $10^{-5}$, momentum of 0.99, a batch size of $100$ and $8$ layers of dilation for SYNTHIA and our dataset and $10$ layers for Richter et al. For Cityscapes we used a learning rate of $10^{-4}$, $10$ layers of dilation, and same momentum and batch size as for synthetic data. A maximum of 40K iterations were used during frontend baseline training, 100K iterations for frontend fine tuning and 60K iterations for context baseline training. 
    
    The project's GitHub page\footnote{https://github.com/fyu/dilation} provides implementation details. Results are given for all classes available in the Cityscapes dataset, except wall, fence and train, which are not present in our dataset. 
    
   \item \phantomsection \label{frrn-a} \textbf{FRRN-A} -- A full-resolution residual network, as described by Pohlen et al.~\cite{pohlen2017FRRN}. The network is trained from scratch with no weight initialization using the same dataset combinations as for DFCN. In this architecture we provide results for all 19 classes available in Cityscapes dataset with IoU scores computed at the native image resolution of the FRRN-A architecture. The reference implementation can be found on  GitHub\footnote{https://github.com/TobyPDE/FRRN}.
   
   The FRRN-A network was trained using ADAM~\cite{Kingma2014} with a learning rate of $10^{-3}$ for organic data and $10^{-4}$ for synthetic data. The bootstrap window size was set to 512 for organic data and 8192 for synthetic data. The batch size was 3. A maximum of 100K iterations were used both during baseline training and fine tuning.
   
\end{enumerate}

\noindent Our primary goal is to explore how the performance of each dataset varies across the different testing conditions and to analyze which properties persist across contexts. In particular, we chose the the DFCN network because of its high performance while still allowing weight initialization from an ImageNet-pretrained VGG architecture. On the other hand, the FRRN network must be trained from scratch, but has the capacity to attain quite high performance on the Cityscapes benchmark. We expect the FRRN architecture to be the most difficult for the synthetic datasets, whereas DFCN's use of the VGG weights may mask domain shift to some degree.

In both architectures, the results given represent the best validation iteration, with snapshots taken at each 2K iterations for DFCN and 1K iterations for FRRN.

\subsection{Results and analysis}

\noindent Table~\ref{table:dfcn_results} presents results of the DFCN front-end module for pure synthetic training as well as versions with fine-tuning on the Cityscapes dataset. On pure synthetic training, our dataset scores 36.93\% with Richter et al. at 31.12\% and SYNTHIA at 20.7\%. Although our method scores highest overall, the gains are largely in the classes on which development was focused, and it is clear that the other classes need further work, e.g. buildings and vegetation, where Richter et al.~contains large variation and performs better. 

In the fine-tuning case, all three datasets improve upon the mean IoU score compared to the Cityscapes baseline. 
Overall, our dataset achieves an IoU of  
69.45\% when fine-tuning on 
Cityscapes. This is in comparison to 
68.83\% for SYNTHIA, and 
69.03\% for Richter et al. 

Table \ref{table:dfcn_results} also gives the scores obtained by the DFCN context module, which acts on the front-end dense predictions in order to increase their accuracy both quantitatively and qualitatively. 
Our dataset scores 39.8\%,  
achieving a further 7.2\% improvement on mean IoU performance compared to the front-end result. In comparison, Richter et al.'s mean IoU score barely improves, and SYNTHIA regresses. We believe this is due to the fact that each of our training images are unique, giving the network more variation at the contextual level. In particular, one of the most important classes -- car -- improves to 73.73\%, achieving the highest score in synthetic only training. 

We note that although all Cityscapes classes are present in SYNTHIA, four achieve zero percent IoU (traffic light, traffic sign, terrain and truck)  in frontend training and three of them (traffic light, terrain and truck) fail to improve in context training.
This is due to the small number of total pixels occupied by these classes, and it highlights the importance of providing enough exemplars in the training dataset for the network to learn from. Likewise, the bicycle class is present in Richter et al., but is not well learned. 

Table~\ref{table:frrn_results} shows results for the FRRN architecture on synthetic-only training as well as fine-tuning. Despite lacking three classes (wall, fence and train), our dataset achieves an IoU of 31.15\%, with SYNTHIA at 21.75\% and Richter et al. at 20.42\%. Notably, the Richter et al. dataset performs worse than SYNTHIA in this architecture, and some classes that perform well with DFCN drop significantly on FRRN, e.g. Richter et al.'s car (73.49\% to 27.30\%) and person (40.56\% to 31.72\%). We attribute this to training the network from scratch, which highlights the domain shift between Richter et al. and Cityscapes. In contrast, our dataset achieves similar scores on both networks (50.97\% to 60.41\% and 49.67\% to 50.67\% for car and person, respectively), indicating less domain shift and less reliance on the initial VGG weights.

Fine tuning on the full set of Cityscapes images yields a score of 71.06\% (a 4.3\% relative increase) for our dataset, with 69.89\% for SYNTHIA and 70.76\% for Richter et al. In the FRRN case it is evident that the network shares its feature weights across classes; 
although our dataset contains no instances of the wall, fence or train classes, with fine-tuning on the Cityscapes dataset the network sees a relative increase in IoU of 10.8\% for wall, 10.1\% for fence and 19.6\% for train. As expected, when a wider range of training data is given for the network to learn class \textbf{X}, class \textbf{Y} can benefit from improvements to the shared set of features. In fact, because those three classes are rarely occurring in the Cityscapes data, the percentage increase in performance is two to four times greater than for more common classes.

When considering the classes on which our development was focused (see Section~\ref{sec:procedural}), we see that we reach the highest score in 7 out of those 8 classes for both the DFCN and FRRN architectures on pure synthetic training. This type of targeting is especially desirable when considering an evolving dataset, and the procedural modeling approach ensures that further variation can be added as needed to specific classes without undoing previous work. 

During training there are oscillations in individual class scores, which stem from the inherent competition between the classes to optimize the shared feature weights to suit each particular class' needs. Table~\ref{table:dfcn_best_class_IoU} shows the difference between the best overall validation iteration (based on mean IoU) and the best per-class IoU across all validation iterations for the DFCN architecture. Table~\ref{table:best_class_iou} shows the same data for FRRN. Here we can see that most classes across all three datasets achieve a result that is higher than the Cityscapes baseline at some point during training, but that the iteration that provides the overall best mean IoU may have several classes performing below their respective optima. For the 16 classes included in our dataset, we achieve the best per-class IoU on 12 classes in DFCN-frontend and 14 classes in FRRN.

There are further conclusions to be drawn from each dataset's performance in the two respective architectures. While the dataset from Richter et al. achieves 50\% higher mean IoU performance than SYNTHIA in the DFCN architecture, it is 6.1\% behind SYNTHIA in the FRRN architecture. We attribute this to the VGG weight initialization used in DFCN, which carries features from pre-training on ImageNet, and which seem to complement features that are lacking in the Richter dataset itself. It is likely the case that low-level features exhibit a large domain shift in Richter et al., but that the greater variation in high-level features due to the extensive game world yields better training. This is further exemplified once fine-tuning is performed on Cityscapes: with the organic dataset added the network pre-trained on Richter et al. again outperforms SYNTHIA.

\begin{figure*}[h]
\rotatebox{90}{SYNTHIA}
  \includegraphics[width=0.32\linewidth]{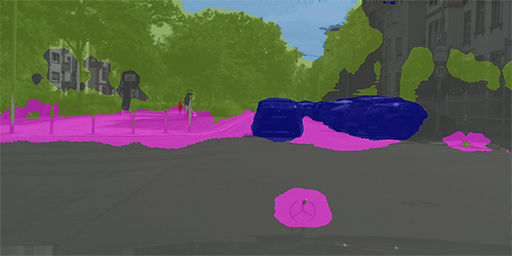} 
  \includegraphics[width=0.32\linewidth]{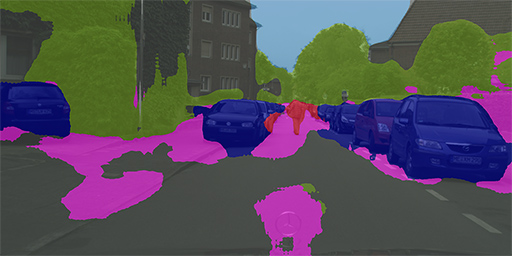} 
  \includegraphics[width=0.32\linewidth]{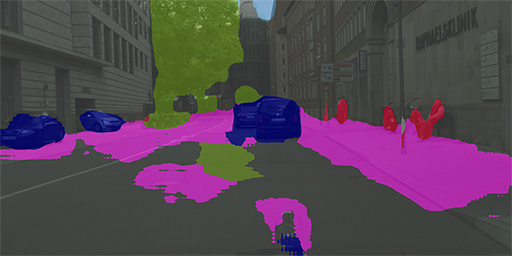} \\ [1pt]
\rotatebox{90}{Richter et al.}
  \includegraphics[width=0.32\linewidth]{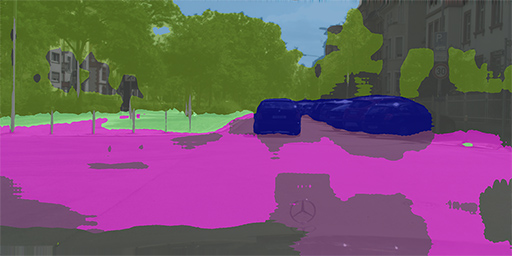} 
  \includegraphics[width=0.32\linewidth]{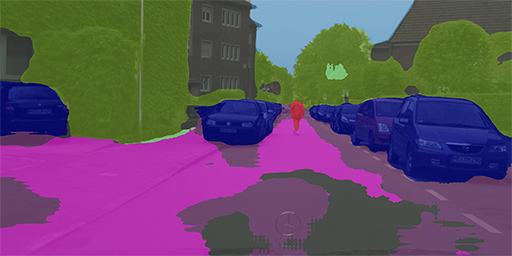} 
  \includegraphics[width=0.32\linewidth]{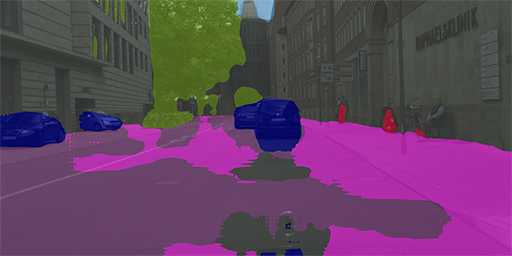} \\ [1pt]
\rotatebox{90}{Our method}
  \includegraphics[width=0.32\linewidth]{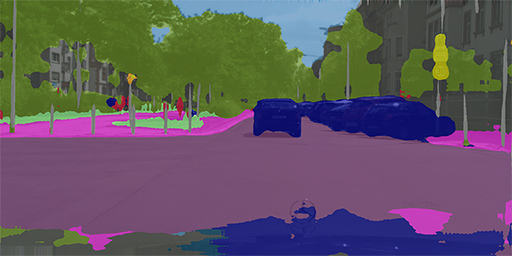} 
  \includegraphics[width=0.32\linewidth]{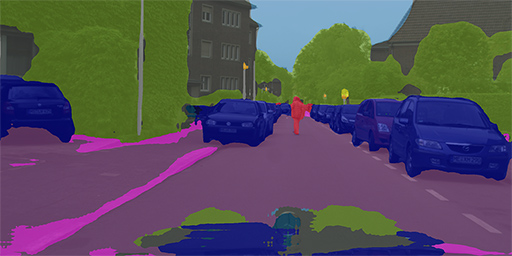} 
  \includegraphics[width=0.32\linewidth]{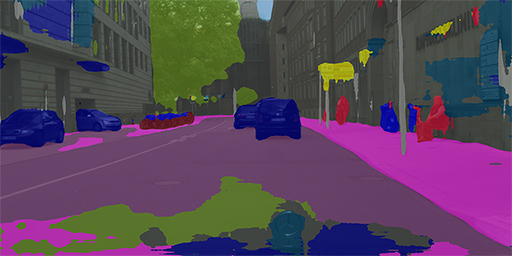}\\ [1pt]
\rotatebox{90}{Ground truth}
  \includegraphics[width=0.32\linewidth]{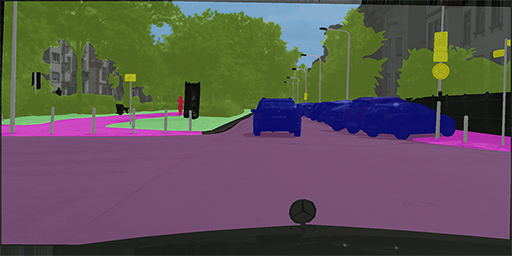}  
  \includegraphics[width=0.32\linewidth]{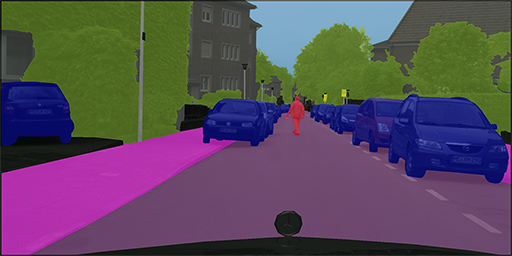}
  \includegraphics[width=0.32\linewidth]{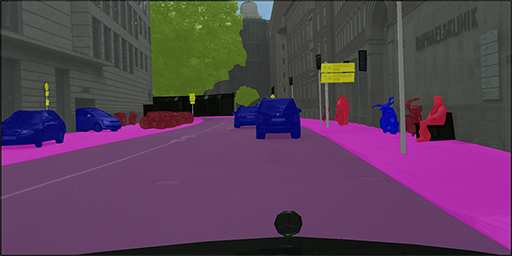} \\ [2pt]
\makelegenddfcn

\caption{
Results for the DFCN front-end architecture on pure synthetic data, corresponding to Table~\ref{table:dfcn_results}. Note the improved road surface and pedestrian segmentation, and the ability to identify traffic signs as well as poles, bicycles and motorcycles with our method.
}
\label{fig:dfcn-frontend_synthetic}
\vspace{-3mm}
\end{figure*}

The three example images in Figure~\ref{fig:dfcn-frontend_synthetic} show the behavior of the DFCN architecture trained on each of the synthetic datasets. The domain shift in the road surface and sidewalk classes is evident in both SYNTHIA and Richter et al., which both suffer from false predictions in large parts of the roadway. Although our dataset performs significantly better on classes such as road, sidewalk, traffic signs, bicycles, poles and motorcycles, it still has problems categorizing buildings correctly, likely due to insufficient variation for that particular class in our training set. 

\begin{figure*}[h]
\rotatebox{90}{SYNTHIA}
\includegraphics[width=0.32\linewidth]{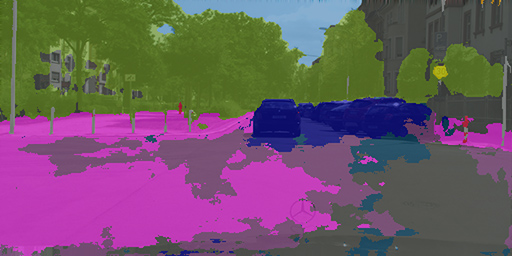}
\includegraphics[width=0.32\linewidth]{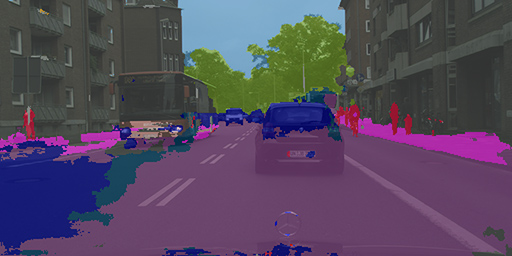}
\includegraphics[width=0.32\linewidth]{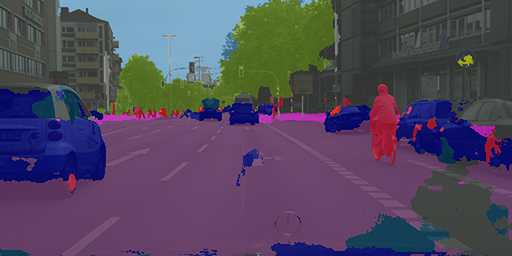}\\ [1pt]
\rotatebox{90}{Richter et al.}
\includegraphics[width=0.32\linewidth]{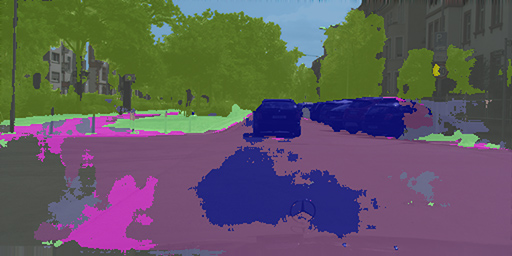}
\includegraphics[width=0.32\linewidth]{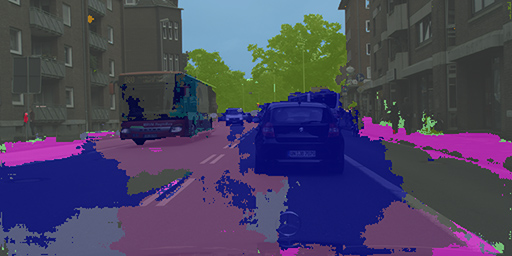}
\includegraphics[width=0.32\linewidth]{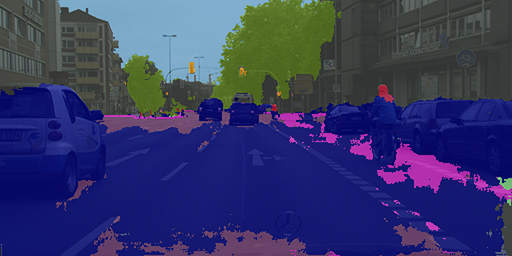}\\ [1pt]
\rotatebox{90}{Our method}
\includegraphics[width=0.32\linewidth]{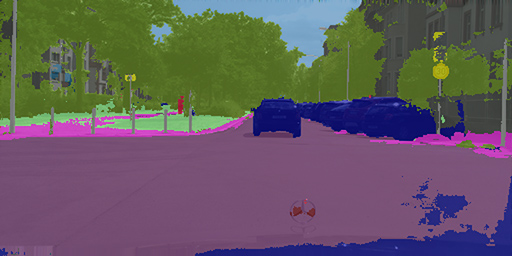}
\includegraphics[width=0.32\linewidth]{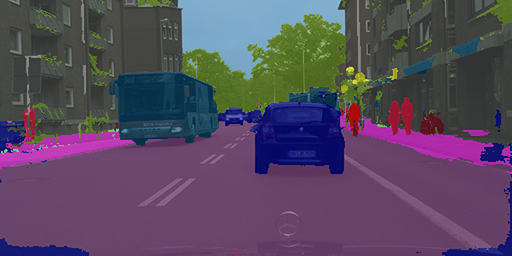}
\includegraphics[width=0.32\linewidth]{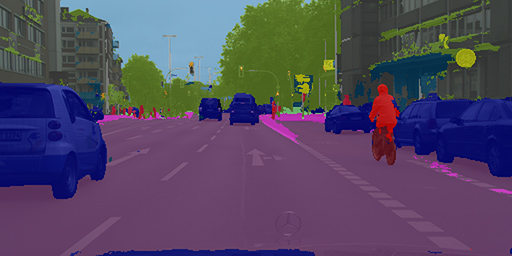}\\ [1pt]
\rotatebox{90}{Ground truth}
\includegraphics[width=0.32\linewidth]{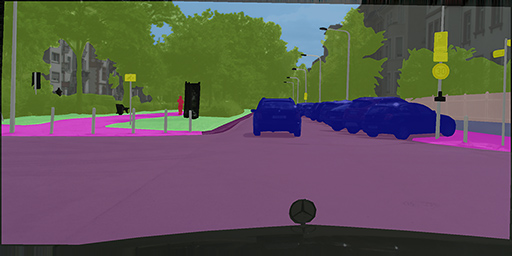}
\includegraphics[width=0.32\linewidth]{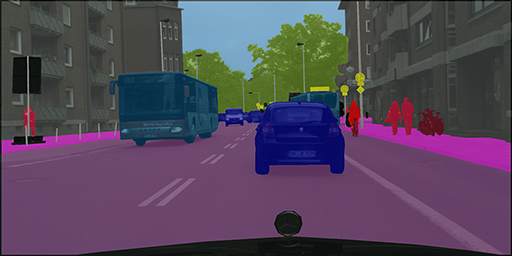}
\includegraphics[width=0.32\linewidth]{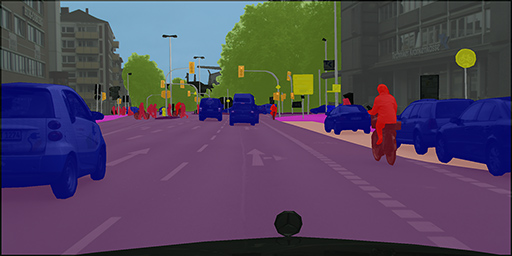} \\ [2pt]
\makelegend

\caption{Results for the FRRN-A architecture, corresponding to Table~\ref{table:frrn_results}. The more severe domain shift in SYNTHIA and Richter et al.~is apparent, and only road, buildings, vegetation, and sky have IoU over 40\%. Our dataset improves performance across nearly all classes, achieving at least 40\% IoU on road, sidewalk, building, sky, person and car.}
\label{fig:frrn_baseline}
\vspace{-3mm}
\end{figure*}

The FRRN examples in Figure~\ref{fig:frrn_baseline} show similar outcomes as DFCN, although the structure of false predictions tends to be of higher frequency due to the residual network architecture. In particular, we see that the network, without weight initialization, is even more susceptible to domain shift than DFCN.

\begin{figure*}[h]
\rotatebox{90}{Cityscapes}
  \includegraphics[width=0.5\linewidth]{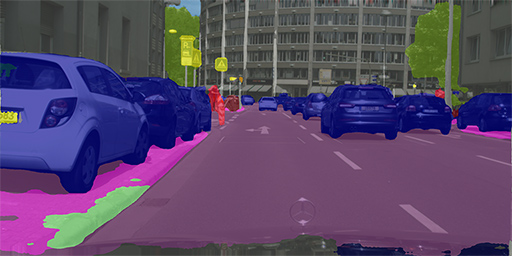}  
  \includegraphics[width=0.5\linewidth]{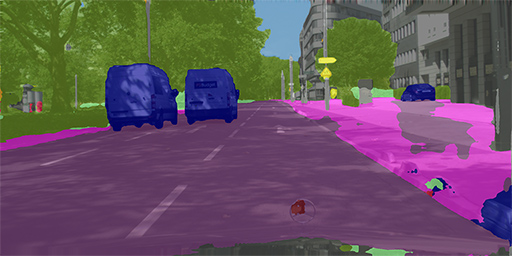} 
\rotatebox{90}{Ours + Cityscapes}
  \includegraphics[width=0.5\linewidth]{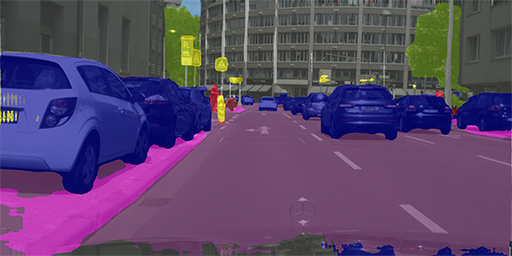}  
  \includegraphics[width=0.5\linewidth]{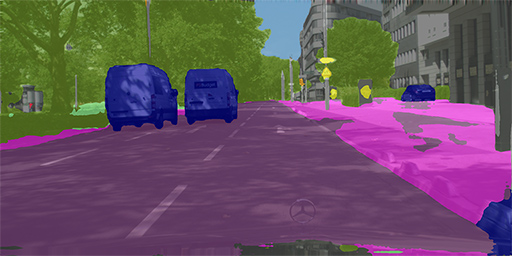} \\ [1pt]
\rotatebox{90}{Cityscapes ground truth} 
\includegraphics[width=0.5\linewidth]{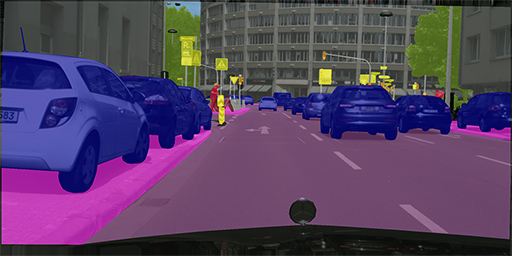}  
  \includegraphics[width=0.5\linewidth]{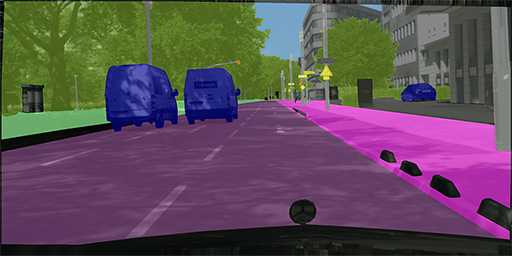} \\ [2pt]
\makelegenddfcn
  
  \caption{DFCN results with fine-tuning. Our dataset helps the network disambiguate between a street sign and a person in the left image, and allows the network to recognize a distant traffic light in the right image.
}
\label{fig:dfcn-frontend_finetune}
\vspace{-3mm}
\end{figure*}

The images in Figure~\ref{fig:dfcn-frontend_finetune} show the results of fine-tuning the DFCN network on the full set of Cityscapes for our dataset in comparison with the predictions obtained by training DFCN on Cityscapes itself.  
Here, the improvements provided by pre-training on our synthetic data act as corrections, such as the vertical traffic sign in the bottom left image being corrected from a misprediction of 'person', and the detection of traffic signs in the middle of the same image, that the Cityscapes-only network fails to catch. 

\begin{figure*}[h]
\rotatebox{90}{Cityscapes}
\includegraphics[width=0.5\linewidth]{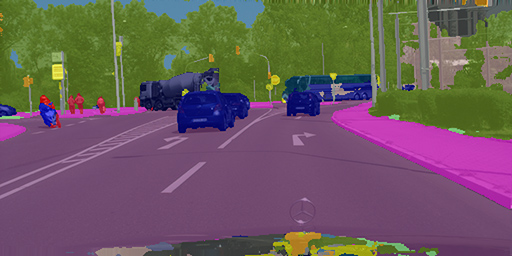}
\includegraphics[width=0.5\linewidth]{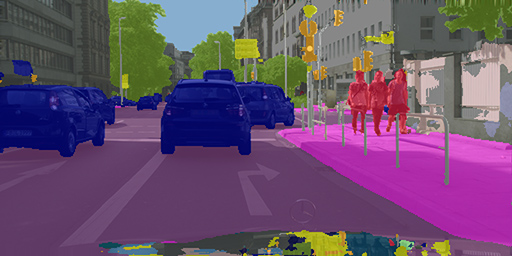} 
\rotatebox{90}{Ours + Cityscapes}
\includegraphics[width=0.5\linewidth]{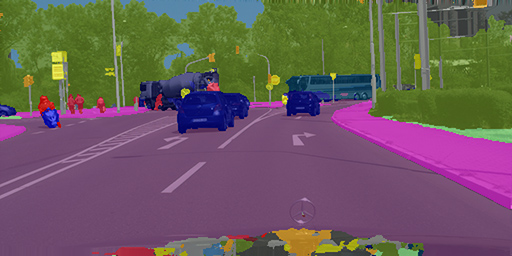}
\includegraphics[width=0.5\linewidth]{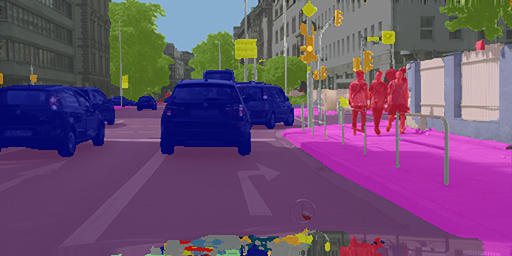}\\ [1pt]
\rotatebox{90}{Cityscapes ground truth} 
\includegraphics[width=0.5\linewidth]{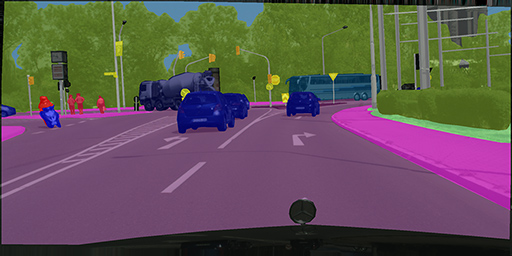}
\includegraphics[width=0.5\linewidth]{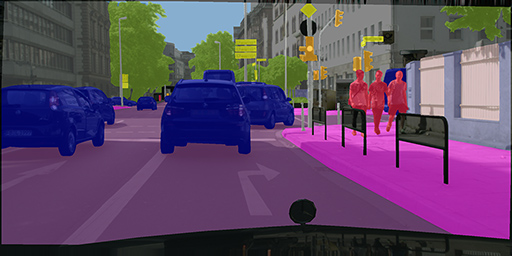}\\ [2pt]
\makelegend

\caption{FRRN-A results with fine-tuning. In the left images we see that our dataset improves the vegetation and terrain recognition as well as that of the distant bus. In the right images we see significant improvement in the wall and fence classes, despite our dataset having no occurrences of either one.}
\label{fig:frrn_finetune}
\vspace{-3mm}
\end{figure*}

In Figure~\ref{fig:frrn_finetune} we see fine-tuning results for the FRRN network. Here, the residual architecture provides more room for silhouette improvements, and we can see tightening of the predictions of 'person' in the bottom right image, as well as examples of the correction to both the wall and fence classes, as discussed previously. We can also see the limitations of judging network performance on hand-annotated data: both the organic-only and the fine-tuned networks correctly predict an additional traffic light on the left side of the image, which is not annotated in the ground truth image.

When considering the training of neural networks,  we would expect a practitioner to choose a training regimen that will yield the highest possible performance, which may include a combination of weight initialization and a mixture of both organic and synthetic data. However, there are many uses for synthetic data besides training: labeled data can also be used for validation of models, for exploration of novel architectures, and for analysis of trained models. In these cases, neither weight initialization nor fine tuning can help bridge the domain shift of datasets with poor realism. GANs have shown some promise in this context, but their use in improving the realism of visual data is so far limited to cases of coarse annotations~\cite{shrivastava2016learning}. At the moment, the best way to produce high quality synthetic data is to engineer the realism into the data itself, rather than attempt to make an unrealistic dataset more realistic through machine learning means. 

\section{Conclusion and future work}
\label{sec:conclusion}
\noindent This paper presented a new approach for generation of synthetic image data with per-pixel accurate annotations for semantic segmentation for training deep learning architectures in computer vision tasks. The image synthesis pipeline is based on procedural world modeling and state-of-the-art light transport simulation using path tracing techniques.

In conclusion, when analyzing the quality of a synthetic dataset, it is in general most telling to perform training on synthetic data alone, without any augmentation in the form of fine-tuning or weight initialization. Our results indicate that differences between datasets at the pure synthetic stage provide the best picture of the relative merits, down to the per-class performance. We also conclude that a focus on maximizing variation and realism is well worth the effort. We estimate that our time investment in creating the dataset is at least three to four orders of magnitude smaller than the much larger virtual world from Richter et al., while still yielding state-of-the-art performance. We accomplish this by ensuring that each image is highly varied as well as realistic, both in terms of low-level features such as anti-aliasing and motion blur, as well as higher-level features where realistic geometric models and light transport comes into play. 

In the future, we will analyze in more detail what impact the realism in the light transport simulation has on the neural network performance, to understand the trade-off between computational cost and inference results. Another interesting venue for future work will be to analyze realism's impact in other important computer vision tasks such as object recognition and feature tracking applications. 

{\small
\bibliographystyle{ieee}
\bibliography{egbib.bib}
}

\end{document}